\documentclass[11pt]{article}
\pdfoutput=1 %

\usepackage[preprint]{acl}

\usepackage{times}
\usepackage{latexsym}
\usepackage[table]{xcolor}
\usepackage[T1]{fontenc}
\usepackage[utf8]{inputenc}
\usepackage{microtype}
\usepackage{inconsolata}
\usepackage{graphicx}
\usepackage{booktabs}
\usepackage{array}
\usepackage{amsmath}
\usepackage{multirow}
\usepackage[table]{xcolor}
\usepackage{pifont}
\usepackage{tikz}
\usepackage{pgfplots}
\usepackage{enumitem}
\usepackage{makecell}
\usepackage{ulem}
\pgfplotsset{compat=1.18}

\usepackage{hyperref}
\definecolor{darkblue}{rgb}{0, 0, 0.5}
\hypersetup{colorlinks=true, citecolor=darkblue, linkcolor=darkblue, urlcolor=darkblue}

\newcommand\blfootnote[1]{%
  \begingroup
  \renewcommand\thefootnote{}\footnote{#1}%
  \addtocounter{footnote}{-1}%
  \endgroup
}

\title{Reinforcement Learning Elicits Contextual \\ Learning of Unseen Language Translation}

\author{
 \textbf{Hanxu Hu\textsuperscript{1}}\quad
 \textbf{Zdeněk Šnajdr\textsuperscript{2}}\quad
 \textbf{Pinzhen Chen\textsuperscript{3}}\quad
 \textbf{Jannis Vamvas\textsuperscript{1}}\quad
 \textbf{Rico Sennrich\textsuperscript{1}}
\\
 \textsuperscript{1}University of Zurich \quad \textsuperscript{2}ETH Zurich\quad \textsuperscript{3}Queen’s University Belfast\\
\texttt{hanxu.hu@uzh.ch}
}

\begin{document}
\maketitle

\begin{abstract}
Prior work has shown that large language models (LLMs) can translate unseen or low-resource languages by undergoing continued training or even by encoding a grammar book in their context. 
However, both methods typically overfit specific languages, with limited zero-shot transfer at test time. 
To translate extremely low-resource languages at scale, we argue that LLMs must acquire the meta-skill of utilizing in-context linguistic knowledge rather than memorizing specific languages. 
In this paper, we propose a reinforcement learning (RL) approach to unseen language translation given rich linguistic context, using a surface-level translation metric (chrF) as the reward. 
Empirically, despite the lightweight reward, our RL-trained models effectively extract and apply relevant linguistic information from the provided context, leading to better translations on completely unseen languages than in-context learning or supervised fine-tuning. 
Our analyses suggest that outcome-based RL can extend beyond conventional reasoning tasks like math and coding to serve as a recipe for language learning from context. \blfootnote{Code at: https://github.com/hanxuhu/rl-new-language}
\end{abstract}

\section{Introduction}
Large Language Models (LLMs) have achieved unprecedented performance across a wide range of tasks. Yet, as ML models, they remain fundamentally constrained when used in data-scarce, out-of-distribution scenarios. A particularly revealing example is the processing of extremely low-resource or entirely unseen languages, e.g. unseen language translation \citep{tanzer2024a}, a setting that we argue constitutes an ideal testbed for LLM research for several reasons. First, reasoning over linguistic patterns with limited pre-training data remains challenging even for frontier LLMs~\citep{bean2024lingoly,kocmi-etal-2025-findings-wmt25}. Second, progress here has direct societal implications: it might enable more effective documentation and preservation of endangered languages, for which large-scale (parallel) corpora hardly exist \citep{joshi-etal-2020-state}.

Two paradigms dominate LLM approaches to new languages: \emph{post-training} the model on additional language/task-specific data \citep{yong-etal-2023-bloom,iyer-etal-2024-exploring}; and where resources are even more limited, \emph{in-context learning} using inductive or deductive examples in the prompt~\citep{garg2022what,agarwal2024many,tanzer2024a}. 
We argue that a more principled and scalable approach should instead be \emph{language-independent learning}, which is inspired by meta-learning \citep{Finn,gu-etal-2018-meta}. Rather than memorize the content or knowledge of any specific language, an LLM should acquire the ability to generalize across new languages by reasoning over an accompanying context with rich linguistic features, grammatical descriptions, morphological paradigms, and lexicons.
We refer to this transferable ability as a \emph{meta-skill of contextual leveraging}, and view it as the key property that any general-purpose method for very low-resource languages should target.

Motivated by this perspective, we introduce a reinforcement learning (RL) method which can elicit LLMs to leverage linguistic context when translating new unseen languages. The recent success of RL with verifiable rewards (RLVR) on reasoning tasks~\citep{openaio1, guo2025deepseek} suggests that outcome-based RL is particularly effective at eliciting transferable skills (e.g. reasoning) rather than memorized fixed solutions. We bring this philosophy to low-resource translation: we treat ``translating a new language given its grammar'' as a verifiable, context-dependent reasoning problem. Concretely, the model is trained to produce translations conditioned on a context of linguistic knowledge and dictionary entries, with a translation quality metric (e.g., chrF) as the outcome reward. By simply rewarding the translation using RL, we can elicit LLMs to learn the meta-skill of leveraging linguistic context and can gain better generalization performance for unseen languages. In summary, our contributions are as follows:
\begin{itemize}[leftmargin=2ex,nosep]
\item \textbf{A reframing of extreme low-resource translation} as the acquisition of a \emph{language-independent meta-skill}, reasoning over in-context linguistic context, rather than memorization of any individual target language, unifying in-context learning and post-training under one objective.
\item \textbf{An outcome-reward RL recipe for context leveraging.} We propose a simple training method that conditions the model on linguistic knowledge and dictionary context and uses translation quality as the outcome reward. 
\item \textbf{Empirical evidence for generalization.} Through controlled context learning comparisons, we show that RL can bring better performance on unseen languages, while SFT overfits to in-domain training languages, showing that the method successfully generalizes.
\end{itemize}

\section{Related Work}

\citet{tanzer2024a} introduced MTOB, a benchmark for translating between English and Kalamang (an extreme low-resource language). By providing a grammar book \citep{eline_visser_2022_6499927} as context, this work demonstrates how LLMs can leverage explicit linguistic documentation to translate an entirely new language. 
\citet{zhang-etal-2024-hire} include bilingual dictionaries, grammar books, and morphologically analyzed text in context to guide LLM translation without parameter updates. 
Following the DiPMT \citep{ghazvininejad2023dictionary} framework, where an LLM is fed bilingual lexicons of rare words in the prompt, \citet{zhang-etal-2024-teaching} used dictionary retrieval and in-context learning to teach LLMs the Zhuang language on the fly. 
\citet{zhang-etal-2025-read} further explored the effectiveness of grammar by decomposing it into rule retrieval and application, and specifically proposed to convert grammar into pseudocode. On the contrary, more recently, \citet{aycock2025can} showed that LLMs make little use of grammar descriptions but rely on the parallel examples in the book when translating an unseen language. This finding is backed by the study by \citet{pei-etal-2025-understanding} on Manchu that bilingual dictionaries are more helpful, while grammar and CoT steps do not bring noticeable benefit. Nonetheless, \citep{marmonier-etal-2025-explicit} found that LLMs have measurable capacity to learn constructed languages from grammar descriptions and fine-tuning on CoT greatly improves it, but generalizability is limited for complex linguistic phenomena. Other approaches using LLMs to process low/no-resource languages include combining an LLM with rules \citep{coleman-etal-2024-llm,coleman-etal-2026-comparing} or finite-state transducers \citep{gutierrez-etal-2025-fsts}.

Orthogonal to above approaches, a recent line of work explores 
fine-tuning via reinforcement learning (RL). Building on RLVR \citep{tulu3,guo2025deepseek} and 
GRPO \citep{deepseekmath}, several works have transferred this paradigm 
to MT despite the absence of a single ``correct'' output. \citet{feng2025mtr1zero} 
introduce MT-R1-Zero, the first R1-Zero-style adaptation for MT, using a hybrid 
BLEU + COMET-Kiwi reward under GRPO; \citet{he2025r1t1fullyincentivizingtranslation} pair COMET-based rewards 
with format signals to incentivize translation reasoning, while 
\citet{wang2025deeptrans} and \citet{yang2026ssr} move beyond reference-based 
metrics via trajectory-level generative or self-rewarding signals. 
\citet{he2024improvingmachinetranslationhuman} earlier showed that quality-estimation models 
provide a data-efficient reward for MT RLHF. Closer to the low-resource regime, 
\citet{mosquera2025improvinglowresourcetranslationdictionaryguided} use RL to teach an LLM 
to consult a bilingual dictionary as an external tool during Spanish–Wayuunaiki 
translation. and \citet{attia2026improvinglowresourcemachinetranslation} propose self-supervised 
round-trip RL with intrinsic chrF++ rewards for Aymara, Friulian, and Wolof. 
However, existing RL-for-MT methods optimize translation quality directly from 
source-target pairs, leaving open how RL can instead teach models to better 
exploit in-context linguistic resources.

Our setup can be naturally framed as meta-learning: rather than optimizing for any single translation instance, we train the model to make better use of a contextual ``support set'' of linguistic resources at inference time. This perspective has a long history in low-resource MT, where \citet{gu-etal-2018-meta} adapted MAML \citep{Finn} to learn initializations that transfer quickly to new language pairs from limited parallel data. With the rise of in-context learning, \citet{brown2020languagemodelsfewshotlearners} reframed few-shot learning as implicit meta-learning over the pre-training distribution, and a subsequent line of work has sought to make this capability explicit: MetaICL \citep{min-etal-2022-metaicl} meta-trains on a distribution of tasks formatted as in-context demonstrations to amplify few-shot performance, in-context tuning \citep{chen-etal-2022-meta} similarly optimizes models to condition on examples rather than to fit them, and \citet{garg2022what} give a theoretical account of what transformers can learn in-context. Unlike these supervised meta-learning approaches, we use RLVR to teach the model to exploit qualitatively heterogeneous in-context resources better for low-resource translation.

\section{Methodology}

\subsection{Data Curation}
\label{sec:data-curation}

\begin{table*}[t]
\centering
\small
\begin{tabular}{lll}
\toprule
\textbf{Component} & \textbf{Content} & \textbf{Retrieval} \\
\midrule
Introduction  & Linguistic and geographic profile of the source language & --- \\
Task instruction    & Translation direction and test sentence& --- \\
Dictionary section  & 2 entries per source token & LCS \\
Parallel sentences  & 3 or 5 source--target pairs& LCS \\
Grammar passages    & 2 raw excerpts from the grammar  & LCS \\
Closing instruction & Request for step-by-step meta-linguistic reasoning & --- \\
\bottomrule
\end{tabular}
\caption{Components of an assembled prompt. The parallel-sentence count (3 or 5) is an experimental condition.}
\label{tab:prompt-components}
\end{table*}

\begin{table*}[t]
  \centering
  \small
  \begin{tabular}{@{}llcccrr@{}}
    \toprule
    \textbf{Split} & \textbf{Group}  & \textbf{Langs} & \textbf{Dirs} & \textbf{Families} & \textbf{Train}     & \textbf{Test}     \\
    \midrule
    \multirow{2}{*}{\textit{Seen}}
      & Romansh\,$\to$\,De (seen varieties)       & 4  & 4$\times$1 & 1  & \textit{13,892} & \textit{1,462} \\
      & Other\,$\leftrightarrow$\,EN            & 7  & 7$\times$2 & 6  &  9{,}695 & ---     \\
    \midrule
    \textit{Similar}
      & Romansh\,$\to$\,De (held-out)$^\ddagger$ & 2  & 2$\times$1 & 1  & \textit{\sout{7,998}} & \textit{737} \\
    \midrule
    \multirow{2}{*}{\textit{Unseen}}
      & Kalamang\,$\leftrightarrow$\,EN$^\dagger$ & 1  & 1$\times$2 & 1  &     \sout{750} &   100   \\
      & OOD (EN\,$\to$\,X only)                 & 4  & 4$\times$1 & 4  & ---      &   400   \\
    \midrule
    \textbf{Total} & &  \textbf{18} & \textbf{26} & \textbf{10} & \textbf{23{,}587} & \textbf{2{,}699} \\
    \bottomrule
  \end{tabular}
  \caption{\textbf{Data summary.}
    Languages are partitioned by evaluation split.
    \textit{Seen} languages appear in both train and test;
    \textit{Similar} directions are held out for evaluation but share their family with seen varieties;
    \textit{Unseen} directions have no related training data.
    \textbf{Only \textit{Seen} languages are used for training};
    struck-through numbers (\sout{7{,}998}, \sout{750}) indicate parallel data that exists for the held-out \textit{Similar} and \textit{Unseen} languages but is \textit{excluded from training}.
    \textbf{Dirs} reports translation directions as $\textit{langs}\times\textit{directionality}$
    ($\times 2$ for bidirectional, $\times 1$ for unidirectional).
    Each prompt contains ${\sim}20$--$34$ dictionary entries, 5 parallel sentence pairs,
    and optionally a grammar-book passage (${\sim}2.8$k tokens on average).
    $^\ddagger$Sursilvan\,$\to$\,De and Surmiran\,$\to$\,De, held out from training.}
  \label{tab:data}
\end{table*}

This section describes the construction of the corpus used for our reinforcement learning experiments, covering language selection, parallel-sentence extraction, the dedicated treatment of Romansh, the synthesis of dictionary entries, and the structure of the assembled prompts.

\subsubsection{Language Selection and Sources}

We investigate fourteen (very) low-resource languages, grouped by data-sourcing pipeline rather than by language family. The first group comprises eight languages whose parallel data is extracted from Language Science Press grammar books: Choguita Rar\'amuri, Gyeli, Japhug, Kagayanen, Kalamang, Tuatschin, Ulwa, and Vamale. We note that Tuatschin is genealogically a dialect of Sursilvan and thus belongs to the Romansh continuum; it is grouped here because its data comes from a grammar book rather than from the parallel corpora and dictionaries used for the standardized Romansh idioms below. The second group consists of the six standardized varieties of Romansh---Puter, Vallader, Surmiran, Sursilvan, Sutsilvan, and Rumantsch Grischun---which, despite a shared heritage, diverge enough lexically and orthographically to warrant separate treatment.

The Romansh varieties were selected because they allow studying meta-learning on a spectrum of language relatedness, and due to the availability of dictionary data and grammar books; other languages were selected based on the availability of open-source grammar books. 
For the non-Romansh languages, candidates were restricted to those with grammars published by Language Science Press under the CC BY 4.0 licence and distributed in both PDF and \LaTeX{} source. The detail is shown in Appendix \ref{app:grammar}.

\subsubsection{Parallel Sentence Extraction}

Due to the lack of parallel corpora available, we extracted translated examples directly from their grammars.
An exception are the five Romansh varieties, where we sample a small amount of parallel sentences from the back-translated training data used by~\citet{vamvas-et-al-2026-translation}.
Although the grammar books occasionally contain OCR or \LaTeX{} artefacts, the underlying translations have been verified by an author-linguist. We worked from the \LaTeX{} source rather than the rendered PDF to preserve idiosyncratic orthographies, locating interlinear-gloss examples by pattern-matching the standard \verb|\gll| and \verb|\glt| commands as well as user-defined macros. Residual markup was stripped from both sides of each pair.

We retained only pairs with source sentences between 6 and 50 space-delimited words. A stricter filter targeting extraction artefacts (long uppercase sequences, inline numbering, stray digits, ellipses, underscores, residual newlines) was tested but reduced the corpora to as few as 11 sentences for Japhug and 20 for Kalamang---too few for reinforcement learning---so only the length filter was retained. To enable bidirectional training, each pair was duplicated with the source and target swapped.

\subsubsection{Romansh Preprocessing}

The Romansh varieties differ in two respects: they possess large, high-quality parallel corpora and dictionaries (grammar is available only for Puter and Vallader), but their reference language is German rather than English. We discarded sentences longer than 50 words from both the training and test splits and, where more than 2{,}000 pairs remained, randomly sampled 2{,}000 to keep per-language volumes comparable. All prompt components were translated into German so that experimental conditions remain identical across language groups.
Dictionary entries where retrieved using the Rumlem tool~\cite{fischer-etal-2026-rumlem} based on the Romansh segments.

\subsubsection{Synthetic Dictionary Augmentation}
\label{sec:synthetic-dict}

Every language other than Kalamang and the Romansh varieties lacks sufficient dictionary resources, so we generated synthetic entries using an LLM based on parallel data and grammar books in the format of the MTOB benchmark \citep{tanzer2024a}. Two prompt designs were compared, conditioning on the same parallel sentences and grammar excerpts but differing in the dictionary section: \textit{v1} explicitly names each source token and shows two working example entries per token, whereas \textit{v2} only describes the general entry format without token-specific demonstrations. For each language and translation direction, we generated dictionaries with both variants, used them in prompts over the training data, and retained the variant with the higher chrF on the resulting translations. The final synthetic dictionary is therefore a per-language, per-direction mixture of v1 and v2.

\subsubsection{Prompt Composition}

Each prompt is assembled from five components, as summarised in Table~\ref{tab:prompt-components}, and ends with an instruction to produce step-by-step meta-linguistic reasoning before committing to a final translation. Retrieval uses the Longest Common Subsequence (LCS) metric, which outperformed embedding-based similarity in preliminary experiments. Two configurations of the parallel-sentence section---three and five examples---are evaluated independently to study the effect of in-context example count.

Table~\ref{tab:data} summarises the composition of our dataset across the four language groups. In total, the corpus spans 18 languages from 10 distinct language families, yielding 32{,}335 training pairs and 2{,}699 test pairs. The Romansh varieties contribute the largest share of the training data (21{,}890 pairs across six varieties translated to and from German), reflecting both the availability of high-quality parallel corpora and the upper bound of 2{,}000 sampled sentences per variety. The remaining seven low-resource languages aligned with English contribute 9{,}695 training pairs but are used exclusively for training, since their grammar book-extracted examples are too scarce to be split further. Kalamang$\leftrightarrow$EN is treated separately: we adopt the original MTOB split of 750 training and 100 test pairs to ensure direct comparability with prior work, and additionally hold it out together with Sursilvan$\leftrightarrow$De and Surmiran$\leftrightarrow$De in our training data configuration to probe cross-lingual generalisation within and across families. Finally, four languages out-of-distribution (one per family) are reserved as a test-only EN $\to$ X benchmark from \citet{hus-anastasopoulos-2024-back} to assess transfer to languages entirely unseen during RL training, with 100 sentences randomly sampled per direction for each language.

\subsection{RL with Linguistic Knowledge in Context}
\label{sec:rl-formulation}

With the curated data from Section~\ref{sec:data-curation}, we formulate translation as a reinforcement learning task in which the policy is trained to produce translations conditioned on a meta-linguistic context.

\paragraph{Task formulation.}
Each training instance is a triple $(x, c, y)$, where $x$ is a source sentence, $y$ its reference translation, and $c$ the language-knowledge context assembled as described in Section~\ref{sec:data-curation}, comprising the introductory description, retrieved dictionary entries, parallel sentences, and grammar passages. The policy $\pi_\theta$ is a large language model that, given the concatenation of $c$ and $x$, generates a response containing step-by-step meta-linguistic reasoning followed by a final hypothesis translation $\hat{y}$. Only $\hat{y}$ is scored against $y$; the reasoning trace is left unconstrained.

\paragraph{Reward.}
We use chrF \citep{popovic-2015-chrf} between $\hat{y}$ and $y$ as the reward signal. Since chrF is reported on a 0--100 scale, we rescale it to $[0, 1]$, which is a typical range for reward functions:
$r(\hat{y}, y) = \frac{1}{100}\mathrm{chrF}(\mathrm{hyp}=\hat{y}, \mathrm{ref}=y)$. 
If the model fails to produce a parsable final translation, the reward is set to zero.

\paragraph{Optimization.}
We optimize $\pi_\theta$ with GRPO \citep{deepseekmath}. For each prompt, $G$ responses $\{\hat{y}_i\}_{i=1}^{G}$ are sampled from the current policy and scored independently. Each reward is then standardized within its group, 
$A_i = \frac{1}{\sigma_G}(r_i - \mu_G)
$, 
where $\mu_G$ and $\sigma_G$ are the mean and standard deviation of $\{r_i\}_{i=1}^{G}$, yielding a group-relative advantage that requires no separate value model. The policy is updated with the standard PPO-style clipped objective using $A_i$, regularized by a KL penalty against a frozen reference policy to limit drift from the initial model. The prompt fed to the policy includes meta-linguistic information about the low-resource language grammar rules, dictionary entries, and parallel sentences, together with the source sentence, allowing the translation quality reward to shape how the policy makes use of these resources when generating  $\hat{y}$.

\begin{table*}[t]
  \centering
  \small
  \setlength{\tabcolsep}{3.5pt}
  \begin{tabular}{@{}ll cc ccccc c@{}}
  \toprule
  & & \multicolumn{2}{c}{\textbf{Romansh}\,$\to$\,\textbf{De}} & \multicolumn{5}{c}{\textbf{Unseen languages} (En\,$\to$\,X)} & \\
  \cmidrule(lr){3-4} \cmidrule(lr){5-9}
  \textbf{Method} & \textbf{Context} & Seen & Similar & Kal & Din & Wol & Gua & Kac & \textbf{Avg.} \\
  \midrule
  \multicolumn{10}{@{}l}{\textbf{Qwen3-4B-Base} experiments} \\
  \midrule
  \rowcolor{gray!15} Qwen3-4B-Base   & \textbf{full} & 0.3413 & 0.3203 & 0.2558 & 0.1606 & 0.1505 & 0.1728 & 0.1774 & 0.2255 \\
  \rowcolor{gray!15} Our SFT          & \textbf{full} & \textbf{0.6017} & \textbf{0.5464} & 0.2860 & 0.0506 & 0.0484 & 0.0639 & 0.0128 & 0.2300 \\
  \rowcolor{gray!15} Our RL           & \textbf{full} & 0.5160 & 0.4785 & \textbf{0.3464} & \textbf{0.2291} & \textbf{0.2253} & \textbf{0.2679} & \textbf{0.2715} & \textbf{0.3335} \\
  \addlinespace[3pt]
  Qwen3-4B-Base   & none & 0.2150 & 0.2413 & 0.1470 & 0.1452 & 0.1303 & 0.1256 & 0.1195 & 0.1606 \\
  Our SFT          & none & 0.4644 & 0.4235 & 0.1358 & 0.1075 & 0.0472 & 0.2083 & 0.1827 & 0.2242 \\
  Our RL           & none & 0.3048 & 0.3694 & 0.1687 & 0.0643 & 0.1025 & 0.1856 & 0.0752 & 0.1815 \\
  \midrule
  \multicolumn{10}{@{}l}{\textbf{Llama-3.2-3B-Instruct} experiments} \\
  \midrule
  \rowcolor{gray!15} Llama-3.2-3B-Inst & \textbf{full} & 0.3012 & 0.2861 & 0.2014 & 0.1044 & 0.1118 & 0.1307 & 0.1545 & 0.1843 \\
  \rowcolor{gray!15} Our SFT          & \textbf{full} & \textbf{0.5577} & \textbf{0.4970} & 0.2444 & 0.0329 & 0.0576 & 0.0713 & 0.0269 & 0.2125 \\
  \rowcolor{gray!15} Our RL           & \textbf{full} & 0.4765 & 0.4414 & \textbf{0.3005} & \textbf{0.1949} & \textbf{0.2029} & \textbf{0.2486} & \textbf{0.2493} & \textbf{0.3020} \\
  \addlinespace[3pt]
  Llama-3.2-3B-Inst & none & 0.2222 & 0.2037 & 0.1485 & 0.0640 & 0.0571 & 0.0784 & 0.0236 & 0.1139 \\
  Our SFT          & none & 0.4411 & 0.3826 & 0.1225 & 0.0410 & 0.1131 & 0.1231 & 0.0152 & 0.1769 \\
  Our RL           & none & 0.3470 & 0.3327 & 0.1669 & 0.0968 & 0.1134 & 0.1577 & 0.0430 & 0.1796 \\
  \bottomrule
  \end{tabular}
  \caption{\textbf{Main results} on low-resource translation (chrF, $\uparrow$).
  \textbf{Seen}: avg of 4 Romansh varieties$\to$German (Puter, Vallader, Sutsilvan, Rumantsch\,Gr.), in
  training data.
  \textbf{Similar}: avg of 2 held-out varieties$\to$German (Sursilvan, Surmiran).
  \textbf{Unseen}: 5 languages from unrelated families, never in training---Kalamang (Papuan), Dinka (Nilo-Saharan),
  Wolof (Niger-Congo), Guarani (Tupi-Guarani), Kachin (Sino-Tibetan).
  \textbf{Avg.}: macro-average across all 7 language directions.
  The \textbf{full-context} block (dictionary $+$ parallel sentences $+$ grammar; shaded) is our primary evaluation; the no-context block (task instruction only) is included to show the expected failure to generalize to unrelated unseen languages.
  \textbf{Bold} = best fine-tuned row per (column) cell within the full-context block.}
  \label{tab:main}
\end{table*}

\section{Experiments}

    \begin{table}[t]
  \centering
  \small
  \setlength{\tabcolsep}{5pt}
  \begin{tabular}{@{}l cc cc@{}}
  \toprule
   & \multicolumn{2}{c}{\textbf{Romansh}\,$\to$\,\textbf{De}} & \multicolumn{2}{c}{\textbf{Kalamang}} \\
  \cmidrule(lr){2-3} \cmidrule(lr){4-5}
  \textbf{Train/Test Context} & \makecell{Puter,\\Val} & \makecell{Surs,\\Surm} & \makecell{En$\to$\\Kal} & \makecell{Kal$\to$\\En} \\
  \midrule
  Full (dict+sent+gram)   & \textbf{.5324} & \textbf{.4785} & \textbf{.3464} & \textbf{.3843} \\
  No grammar (dict+sent)  & .5249    & .4766    & .3319    & .3821 \\
  No sent (dict+gram)& .5224    & .4669    & .2733    & .2932 \\
  No dict (sent+gram)& .4483   & .4077    & .2626    & .3146 \\
  Task only (no context)     & .4154    & .3770    & .1700    & .1968 \\
  \bottomrule
  \end{tabular}
  \caption{\textbf{Context ablation} (RL, Qwen3-4B-Base). Each row trains and evaluates with a matched
  context level. Dictionary entries are the most impactful component ($-$8 chrF on seen languages when removed); grammar
  contributes minimally ($-$0.5); parallel sentences are critical for OOD Kalamang ($-$7 on En$\to$Kal).}
  \label{tab:context-ablation}
  \end{table}

\subsection{Experimental Setup}
\label{sec:exp-setup}

\paragraph{Models and training.}
We fine-tune two backbones, \textbf{Qwen3-4B-Base} and \textbf{Llama-3.2-3B-Instruct}, with both SFT and RL. SFT and RL share an identical prompt format---the full retrieval context from Section~\ref{sec:data-curation}, with two LCS-retrieved dictionary entries per source token, three or five parallel sentences, and two grammar passages where available---and differ only in the supervision signal: SFT minimises cross-entropy against the gold reference, whereas RL optimises chrF reward via GRPO, as described in Section~\ref{sec:rl-formulation}. All models are trained for one epoch on the 22 training directions, which exclude the two Romansh varieties (Sursilvan\textrightarrow{}German, Surmiran\textrightarrow{}German) held out for the \textit{Similar} evaluation setting. SFT uses an effective batch size of 128 and RL a batch size of 64. Remaining hyperparameters are listed in Appendix~\ref{app:hparams}.

\paragraph{Evaluation data.}
For Romansh–German evaluation, we use the WMT24++ benchmark~\cite{vamvas-etal-2025-expanding}, with each language variety evaluated in the X$\to$De direction.
For the five unseen languages, the test data come from two sources. Kalamang (100 pairs in each direction) uses the original MTOB benchmark split \citep{tanzer2024a}, based on the Kalamang grammar book \citep{eline_visser_2022_6499927}. The remaining four languages (Dinka, Wolof, Guarani, Kachin, 100 pairs each in the EN$\to$X direction) use the test sets from \citet{hus-anastasopoulos-2024-back}.

\paragraph{Test-time conditions and metric.}
Each checkpoint is evaluated under two prompt {context} conditions to separate retrieval-dependent skills from those internalized into the policy: \textbf{full} (same prompt as training) and \textbf{none} (only the language description and translation instruction). No model is trained without retrieval. We report chrF results on the 0--1 scale.

\begin{figure*}[t]
  \centering
  \includegraphics[trim={0 0 0 4.25cm},clip,width=\textwidth]{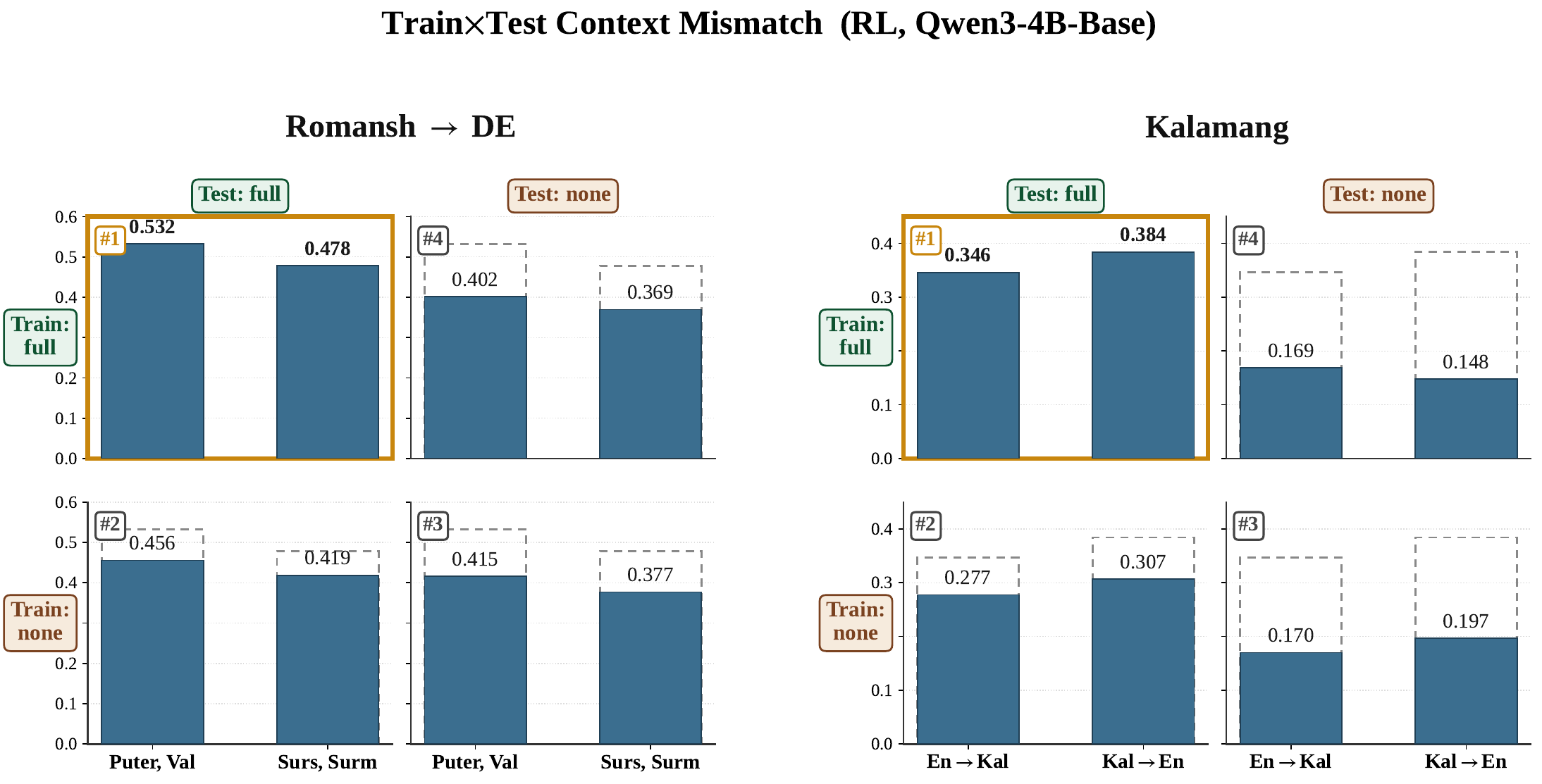}
  \vspace{-1ex}
  \begin{minipage}[t]{0.48\textwidth}
    \centering
    \small Romansh\,$\to$\,De
  \end{minipage}\hfill
  \begin{minipage}[t]{0.48\textwidth}
    \centering
    \small En$\leftrightarrow$Kal
  \end{minipage}
  \caption{\textbf{Train--test context mismatch} (RL, Qwen3-4B-Base). Test-time context dominates: \texttt{no/full} > \texttt{full/no} in every panel (En$\to$Kal: $0.28$ vs.\ $0.17$).}
  \label{fig:ctx-mismatch}
\end{figure*}

\subsection{Comparison of SFT and RL}
\label{sec:rl-generalization}

As shown in Table \ref{tab:main}, we evaluate three conditions of increasing difficulty: (1)~\emph{seen languages} Romansh language varieties seen during training, (2)~\emph{similar} held-out Romansh varieties from the same family, and (3)~five \emph{unseen} languages from unrelated families.
Table~\ref{tab:main} reports results under both full retrieval context and no context (task instruction only).

\paragraph{SFT is stronger on seen languages.}
On seen Romansh$\to$German (averaged over four trained language varieties), Qwen SFT scores 0.60 versus RL's 0.52; LLaMA SFT scores 0.56 versus RL's 0.48.
A similar gap holds on the held-out Romansh varieties (Qwen SFT 0.55 vs.\ RL 0.48), suggesting that SFT's advantage extends within the language family.
This is expected: SFT trains on gold translations to fit word-level alignment, whereas RL receives only a sentence-level chrF reward.

\paragraph{RL is stronger on unseen languages.}
On five unseen languages from different families (Kalamang, Dinka, Wolof, Guarani, Kachin), the ranking reverses.
Qwen RL averages 0.27 across these languages, compared to 0.09 for SFT and 0.18 for the untuned base model.
Llama-3.2 shows a consistent pattern (RL 0.24 vs.\ SFT 0.09 vs.\ base 0.14).
The per-language columns in Table~\ref{tab:main} show that RL outperforms SFT on each of the five languages individually.
Notably, SFT scores below the base model on these languages, suggesting that supervised training on the Romansh directions may reduce the model's ability to use in-context resources for unfamiliar languages.

\paragraph{Context removal clarifies the source of each method's strength.}
Stripping retrieval context at test time (``no context'' rows) reveals different behaviours:
\begin{itemize}[leftmargin=2ex,nosep]
\item \textbf{Seen:} SFT retains much of its performance without context (Qwen SFT: 0.60$\to$0.46), indicating that it stores language-specific mappings in the weights. RL drops more (0.52$\to$0.30), consistent with greater reliance on context.

\item \textbf{Unseen languages:} Without context, all models perform significantly worse.

\item \textbf{The role of context:} RL's advantage on unseen languages (0.27 vs.\ 0.12) appears only when retrieval context is available. This suggests that RL's generalization is tied to its use of in-context resources rather than to language-specific knowledge acquired during training.
\end{itemize}
Taken together, the results point to a trade-off between the two training objectives.
SFT fits the training languages more tightly but appears to reduce context utilization for new languages.
RL fits training languages less precisely but develops a more transferable ability to translate through the provided context: dictionary entries, parallel sentences, and grammar passages, which generalizes to languages from unseen families when such resources are available at test time.

\subsection{Ablation study with each role in context}
\label{sec:context-ablation}

To understand the contribution of each retrieval component, we train five matched RL runs on Qwen3-4B-Base. In each run, we remove one component from both training and test prompts, so the policy never sees a component at training that will be absent at inference. Results are shown in Table~\ref{tab:context-ablation}.

Among the three components, the bilingual dictionary has the largest impact. Removing it causes a drop of 8.4 chrF on seen Romansh (Puter/Vallader: $0.5324 \to 0.4483$) and 8.4 on En$\to$Kalamang ($0.3464 \to 0.2626$). This is expected because the dictionary provides direct word-level grounding that the other two components cannot fully replace.

Parallel sentences rank second. Removing them has only a moderate effect on seen Romansh ($-$1.0 chrF on Puter/Vallader), but a much larger effect on OOD Kalamang ($-$7.3 on En$\to$Kal, from $0.3464$ to $0.2733$). This gap suggests that for languages far from the training distribution, parallel examples provide a complete example for translation.

Grammar passages contribute the least. Removing them causes the smallest drop across all test sets ($-$0.8 chrF on Puter/Vallader, $-$1.5 on En$\to$Kal). The model might be unable to extract useful information from grammar passages.

We also examine whether the benefit of context comes from training or from test time (Figure~\ref{fig:ctx-mismatch}). We cross the two conditions: a model trained \emph{without} context but evaluated \emph{with} context (\texttt{no/full}), versus a model trained \emph{with} context but evaluated \emph{without} (\texttt{full/no}). Across all test sets, \texttt{no/full} consistently outperforms \texttt{full/no} (e.g., En$\to$Kal: $0.28$ vs.\ $0.17$). This shows that test-time context availability is the dominant factor: even a model trained with RL that never saw retrieval during training can exploit it at inference. However, training with context further amplifies this ability. \texttt{full/full} exceeds \texttt{no/full} by $+$7 chrF on En$\to$Kal ($0.35$ vs.\ $0.28$), confirming that exposure to retrieval during RL teaches the policy a more systematic way to use in-context evidence.

\subsection{Analysis of training rewards}
\label{sec:rl-prompt-ablation}

\begin{figure*}[t]
    \centering
    \includegraphics[width=0.85\linewidth]{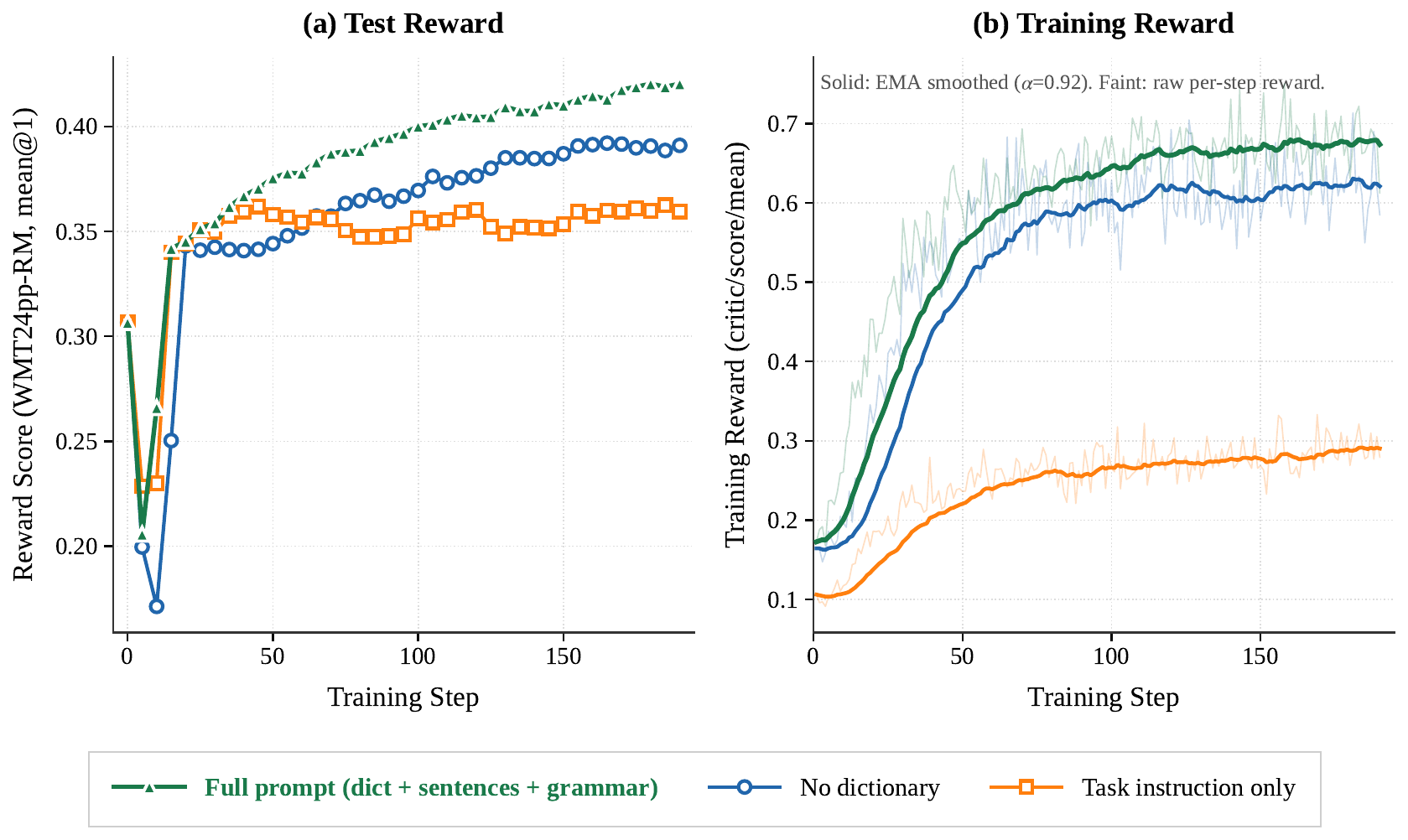}
    \caption{\textbf{RL reward trajectories on Qwen3-4B-Base under three prompt configurations.} (a)~Held-out WMT24++ reward. (b)~chrF training reward; faint lines raw, solid lines EMA-smoothed ($\alpha{=}0.92$).}
    \label{fig:rl-prompt-ablation}
\end{figure*}

\begin{table*}[t]
\centering
\small
\setlength{\tabcolsep}{4pt}
\renewcommand{\arraystretch}{1.15}
\newcolumntype{L}[1]{>{\raggedright\arraybackslash}p{#1}}
\begin{tabular}{@{}l L{0.24\textwidth} L{0.22\textwidth} L{0.22\textwidth} L{0.22\textwidth}@{}}
\toprule
 & \textbf{No context SFT} & \textbf{No context RL} & \textbf{Full context SFT} & \textbf{Full context RL} \\
\midrule
\multicolumn{5}{@{}l@{}}{\textbf{(1) Reference:} Granny Ruslan's grandmother is still strong.} \\
\addlinespace[2pt]
 & Nina Ruslan is a strong woman.
 & Nina Ruslan speaks a strong word.
 & Granny Ruslan is still strong.
 & Granny Ruslan's grandmother is still strong. \\
\midrule
\multicolumn{5}{@{}l@{}}{\textbf{(2) Reference:} Rustam wanted to bathe but he felt cold.} \\
\addlinespace[2pt]
 & Rustam and Waruotkin are the ones who are eating the food.
 & Rustam sees and hears the sound.
 & Rustam wants to wash his hands.
 & Rustam wants to wash and he feels cold. \\
\bottomrule
\end{tabular}
\caption{\textbf{Kalamang\,$\to$\,English case study.} Outputs of SFT and RL models with and without retrieval context, for two source sentences.}
\label{tab:case-study}
\end{table*}

We run GRPO three times on Qwen3-4B-Base over the training split, varying only the retrieval content of the prompt. \textbf{Full} matches Section~\ref{sec:exp-setup} (dictionary, parallel sentences, and grammar passages, all LCS-retrieved). \textbf{No-dict} drops the dictionary block. \textbf{Task-only} drops all three components. The reward function, optimiser, and training budget are unchanged across runs.

Both panels of Figure~\ref{fig:rl-prompt-ablation} rank the three runs in the same order: Full $>$ No-dict $>$ Task-only. With the chrF reward, Full reaches 0.68, No-dict reaches 0.62, and Task-only flattens near 0.29 by step 50. The 6-point gap between Full and No-dict isolates the dictionary's contribution, which is not redundant with parallel sentences. The 33-point gap between No-dict and Task-only shows that without any retrieval, GRPO has nothing language-specific to ground on and saturates near the untuned baseline. On WMT24++, the same order holds, but the differences compress to about 0.06, since the RM weighs fluency more heavily than surface form. 

The reward trajectories support the ablation findings from the perspective of training dynamics. The Task-only run plateaus early, confirming that without linguistic grounding, GRPO exhausts its improvement budget within about 50 steps and converges to a policy that relies only on pretraining priors. The No-dict run continues to improve past step 100, showing that parallel sentences alone provide a useful learning signal, but the ceiling is lower. Only the Full run sustains reward growth throughout training. This indicates that the dictionary supplies a complementary gradient signal that keeps the policy improving even after the parallel-sentence signal saturates. Together with the ablation results, the reward analysis shows that each retrieval component contributes an additive learning signal, with the dictionary providing the strongest boost and grammar the weakest.

\subsection{Case study}
Table~\ref{tab:case-study} illustrates this gap with two examples. 
Without retrieved context, both SFT and RL produce fluent English that recognizes the person name but is semantically unrelated to the source. With a dictionary and parallel-sentence context, RL produces near-perfect translations: in~(1) it exactly matches the reference, and in~(2) it correctly captures both ``bathe'' and ``cold''. SFT, by contrast, captures only partial meaning---rendering ``bathe'' as ``wash his
hands'' and dropping ``cold'' entirely.

\section{Conclusion}
In this paper, we propose a reinforcement learning approach for low-resource translation that learns to exploit in-context linguistic information. Using translation quality directly as the reward signal, our empirical results demonstrate that RL-trained LLMs generalize substantially better to unseen languages than their SFT counterparts, and genuinely learn to leverage linguistic context rather than memorize the specific languages seen during training. Our work offers a new perspective on low-resource translation by uniting the complementary strengths of in-context learning and reinforcement learning.

\section*{Limitations}
We report chrF++ but do not conduct human evaluation, which would give a sharper picture of fluency and adequacy in low-resource settings; we view the automatic numbers as a reliable signal of relative improvement across methods and leave human evaluation to future work. While our method generalizes to unseen languages far better than SFT, absolute performance on unseen languages still lags behind that for higher-resourced languages, indicating a clear headroom for richer in-context evidence and stronger context-utilization signals.

\section*{Acknowledgments}
HH, JV and RS acknowledge funding by the Swiss National Science Foundation (project \mbox{InvestigaDiff}; no.~10000503).
We thank RTR and Fundaziun Patrimoni Cultural RTR for their support, and Lia Rumantscha and Uniun dals Grischs for contributing dictionary data.

\bibliography{bibs/anthology.min,bibs/custom}

\appendix
\section{Hyperparameters}
We perform full-parameter fine-tuning with DeepSpeed ZeRO-2~\cite{rajbhandari2020zero}, using a learning rate of $1\mathrm{e}{-5}$ with a cosine scheduler and 10\% warmup, an effective batch size of 128 (4 per device $\times$ 8 gradient accumulation steps). For RL, we use GRPO with chrF/100 as the outcome reward, a learning rate of $1\mathrm{e}{-6}$, batch size 64, $n=8$ rollouts per prompt at temperature 1.0, clip ratio 0.25, KL coefficient $1\mathrm{e}{-4}$, and entropy coefficient $1\mathrm{e}{-3}$; training uses FSDP with SGLang (TP=4) for rollout and all ablation variants (\textit{nodict}, \textit{nogram}, \textit{nosent}, \textit{taskonly}) share the same hyperparameters.
\label{app:hparams}

\section{LLM Usage Declaration}
We used LLM in this research for two purposes. First, during manuscript writing, we used AI to edit language and polish the overall text. Second, we used GPT-5 mini to generate synthetic bilingual dictionary entries for languages lacking sufficient dictionary resources, as described in Section~\ref{sec:synthetic-dict}. Since none of the authors is a speaker of the target languages, these synthetic entries were not manually validated for linguistic correctness; instead, we performed an indirect, utility-based selection between two prompt variants (v1 and v2) by comparing downstream chrF on the resulting translations and retaining the better-performing variant per language and direction.

We declare that ideation, methodology, experiment execution, and analysis are the original work of the authors. All AI-assisted text editing has been carefully reviewed by the authors to ensure accuracy.

\section{Grammar Books Used For Data Curation}
\label{app:grammar}
\begin{table}[h]
  \centering
  \small
  \begin{tabular}{@{}ll@{}}
    \toprule
    \textbf{Language} & \textbf{Grammar} \\
    \midrule
    \multicolumn{2}{@{}l}{\textit{Open-access grammars (Language Science Press)}} \\
    Choguita Rarámuri & \citet{Caballero2022} \\
    Gyeli             & \citet{Grimm2021} \\
    Japhug            & \citet{Jacques2025} \\
    Kagayanen         & \citet{Pebley2024} \\
    Tuatschin         & \citet{Maurer-Cecchini2021} \\
    Ulwa (PNG)        & \citet{Barlow2023} \\
    Vamale            & \citet{Rohleder2024} \\
    \midrule
    \multicolumn{2}{@{}l}{\textit{Romansh idiom grammars (Lia Rumantscha, print)}} \\
    Surmiran          & \citet{thoeny1969surmiran} \\
    Puter             & \citet{ganzoni1983puter} \\
    \bottomrule
  \end{tabular}
  \caption{Grammar-book sources used for data curation (\S\ref{sec:data-curation}). The eight grammars in the upper block are published by Language Science Press under the CC BY 4.0 license with \LaTeX{} source available, from which we extracted parallel translation examples. The two Romansh idiom grammars in the lower block are published by Lia Rumantscha as print volumes; for these we extracted examples from the printed text rather than \LaTeX{} source.}
  \label{tab:grammar-sources}
\end{table}

\providecommand{\cmark}{\ding{51}}
\providecommand{\xmark}{\ding{55}}
\providecommand{\synth}{$\triangle$} %

\providecommand{\cmark}{\ding{51}}
\providecommand{\xmark}{\ding{55}}
\providecommand{\synth}{$\triangle$} %

\providecommand{\cmark}{\ding{51}}
\providecommand{\xmark}{\ding{55}}
\providecommand{\synth}{$\triangle$} %

\begin{table*}[t]
  \centering
  \small
  \setlength{\tabcolsep}{4pt}
  \renewcommand{\arraystretch}{1.15}
  \begin{tabular}{@{}ll cc cc l r@{}}
    \toprule
    \textbf{Language} & \textbf{Family} & \textbf{Split} & \textbf{Dir.}
      & \textbf{Dict.} & \textbf{Par.} & \textbf{Grammar} & \textbf{Pairs} \\
    \midrule
    \multicolumn{8}{@{}l}{\textit{\textbf{Romansh continuum} (ref.\ language: German)}} \\
    \addlinespace[1pt]
    Puter         & Romance & Seen    & $\to$De & \cmark & \cmark & \citealp{ganzoni1983puter} & \multirow{4}{*}{\makecell[r]{13{,}892\,/\\1{,}462}} \\
    Vallader      & Romance & Seen    & $\to$De & \cmark & \cmark & \xmark & \\
    Sutsilvan     & Romance & Seen    & $\to$De & \cmark & \cmark & \xmark & \\
    Rumantsch Gr. & Romance & Seen    & $\to$De & \cmark & \cmark & \xmark & \\
    \addlinespace[1pt]
    Sursilvan     & Romance & Similar & $\to$De & \cmark & \cmark & \xmark & \multirow{2}{*}{\makecell[r]{\sout{7{,}998}\,/\\737}} \\
    Surmiran      & Romance & Similar & $\to$De & \cmark & \cmark & \citealp{thoeny1969surmiran} & \\
    \midrule
    \multicolumn{8}{@{}l}{\textit{\textbf{Grammar-book languages} (parallel data from LSP grammars; ref.\ language: English)}} \\
    \addlinespace[1pt]
    Choguita Rar\'amuri & Uto-Aztecan  & Seen & $\leftrightarrow$En & \synth & \cmark & \citealp{Caballero2022}        & \multirow{7}{*}{\makecell[r]{9{,}695\\(7 langs)}} \\
    Gyeli               & Niger-Congo  & Seen & $\leftrightarrow$En & \synth & \cmark & \citealp{Grimm2021}            & \\
    Japhug              & Sino-Tibetan & Seen & $\leftrightarrow$En & \synth & \cmark & \citealp{Jacques2025}          & \\
    Kagayanen           & Austronesian & Seen & $\leftrightarrow$En & \synth & \cmark & \citealp{Pebley2024}           & \\
    Tuatschin           & Romance      & Seen & $\leftrightarrow$En & \synth & \cmark & \citealp{Maurer-Cecchini2021}  & \\
    Ulwa (PNG)          & Keram        & Seen & $\leftrightarrow$En & \synth & \cmark & \citealp{Barlow2023}           & \\
    Vamale              & Austronesian & Seen & $\leftrightarrow$En & \synth & \cmark & \citealp{Rohleder2024}         & \\
    \midrule
    \multicolumn{8}{@{}l}{\textit{\textbf{Unseen / OOD languages} (test-only; EN$\to$X sets from \citealp{hus-anastasopoulos-2024-back})}} \\
    \addlinespace[1pt]
    Kalamang  & Gr.\ W.\ Bomberai & Unseen & $\leftrightarrow$En & \cmark & \cmark & \citealp{tanzer2024a} & \sout{750}\,/\,100 \\
    Dinka     & Nilo-Saharan      & Unseen & En$\to$X & \cmark & \cmark &  & 100 \\
    Wolof     & Niger-Congo       & Unseen & En$\to$X & \cmark & \cmark &  & 100 \\
    Guarani   & Tupi-Guarani      & Unseen & En$\to$X & \cmark & \cmark &  & 100 \\
    Kachin    & Sino-Tibetan      & Unseen & En$\to$X & \cmark & \cmark & & 100 \\
    \bottomrule
  \end{tabular}
  \caption{\textbf{Per-language resources and usage.}
    For each language we list its \textbf{Family}, evaluation \textbf{Split}
    (\textit{Seen}: train$+$test; \textit{Similar}: held-out variety of a seen family;
    \textit{Unseen}: no related training data), translation \textbf{Dir.}ection,
    the in-context resources used---bilingual \textbf{Dict.}ionary and \textbf{Par.}allel sentences
    (\cmark\ curated, \synth\ LLM-synthesised, \xmark\ none)---and the \textbf{Grammar} book source
    (citation if used, \xmark\ otherwise; see Table~\ref{tab:grammar-sources}).
    \textbf{Pairs} gives train\,/\,test counts; Romansh and the 7-language group report group totals.
    \sout{Struck-through} counts exist but are \emph{excluded from RL training}.
    Romansh varieties use German as the reference language; all others use English.
    Kalamang uses the MTOB split \citep{tanzer2024a}; the four OOD EN$\to$X test sets
    are from \citet{hus-anastasopoulos-2024-back}.}
  \label{tab:per-language-details}
\end{table*}

\begin{table*}[t]
\centering
\small
\caption{Structure of the translation prompt for Romansh$\rightarrow$German. The prompt consists of a linguistic introduction, a translation instruction, retrieved dictionary entries, parallel sentence examples, a grammar excerpt, and a closing instruction. Placeholder text is shown in \textit{italics}. Dictionary entries are available for Puter and Vallader.}
\label{tab:prompt_structure_romansh}
\renewcommand{\arraystretch}{1.3}
\begin{tabular}{@{}p{2.8cm}p{5.2cm}p{6.5cm}@{}}
\toprule
\textbf{Component} & \textbf{Description} & \textbf{Template (German)} \\
\midrule

\textsc{Linguistic Introduction} &
Background paragraph introducing the Romansh language family, its status as a Swiss minority language in Graubünden, and its five written varieties plus Rumantsch Grischun. &
\textit{R\"atoromanisch geh\"ort zum romanischen Zweig der indogermanischen Sprachfamilie. Es ist eine Minderheitensprache im Schweizer Kanton Graub\"unden \ldots} \\

\midrule
\textsc{Translation Instruction} &
Specifies the source variety, target language, and the sentence to be translated. &
\"Ubersetze von \textit{[Variety]} nach Deutsch: \textit{[source sentence]} \\

\midrule
\textsc{Dictionary Entries} \newline ($\leq$100 entries) &
Retrieved bilingual dictionary entries for words in the source sentence. Each entry provides the closest match from a variety-specific dictionary. &
Um bei der \"Ubersetzung zu helfen, hier ist einer der \"ahnlichsten Eintr\"age zu ``\textit{[word]}'' im \textit{[Variety]}-Deutschen W\"orterbuch:\newline \textit{[Variety]}: \textit{[source term]}\newline Deutsch: \textit{[target term]} \\

\midrule
\textsc{Parallel Sentences} \newline ($\leq$5 pairs) &
Retrieved parallel sentence pairs containing words similar to those in the source sentence. Provides in-context translation examples. &
Um bei der \"Ubersetzung zu helfen, hier ist ein \"ubersetzter Satz mit W\"ortern \"ahnlich zu ``\textit{[word]}'' in einer Liste \"ubersetzter \textit{[Variety]}-Deutscher Referenzs\"atze:\newline \textit{[Variety]}: \textit{[source sentence]}\newline Deutsch: \textit{[target sentence]} \\

\midrule
\textsc{Grammar Passage} &
An excerpt from a grammar book of the source variety, providing morphological and phonological rules. &
Um bei der \"Ubersetzung zu helfen, hier ein aus einem \textit{[Variety]} Grammatiksbuch entnommener Abschnitt:\newline \textit{[grammar excerpt]} \\

\midrule
\textsc{Closing Instruction} &
Instructs the model to provide step-by-step metalinguistic reasoning and place the final translation in a \texttt{\textbackslash boxed\{\}} environment. &
Schreibe nun die \"Ubersetzung des Satzes: \textit{[source sentence]}\newline Bitte lege deine metalinguistischen \"Uberlegungen Schritt f\"ur Schritt dar und setze deine endg\"ultige \"Ubersetzung in eine Box: \texttt{\textbackslash boxed\{\}} \\

\bottomrule
\end{tabular}
\end{table*}

\begin{table*}[t]
\centering
\small
\caption{Structure of the translation prompt for Kalamang$\rightarrow$English. The prompt includes dictionary entries, parallel sentence examples, and grammar passages. The same structure applies to other endangered languages (Gyeli, Japhung, Ulwa, etc.) where a bilingual dictionary is available. Placeholder text is shown in \textit{italics}.}
\label{tab:prompt_structure_kalamang}
\renewcommand{\arraystretch}{1.3}
\begin{tabular}{@{}p{2.8cm}p{5.2cm}p{6.5cm}@{}}
\toprule
\textbf{Component} & \textbf{Description} & \textbf{Template (English)} \\
\midrule

\textsc{Linguistic Introduction} &
Brief description of the language family and geographic location of the source language. &
\textit{Kalamang is a Papuan language of the Greater West Bomberai family. It is spoken in East Indonesia.} \\

\midrule
\textsc{Translation Instruction} &
Specifies the source language, target language, and the sentence to be translated. &
Your task is to translate from \textit{[Language]} to English: \textit{[source sentence]} \\

\midrule
\textsc{Dictionary Entries} &
Retrieved entries from a bilingual dictionary for words in the source sentence. Each entry provides the closest match with part of speech and English translation. &
To help with the translation, here is one of the closest entries to ``\textit{[word]}'' in the \textit{[Language]}-English bilingual dictionary:\newline \textit{[Language]} word: \textit{[source term]}\newline Part of speech: \textit{[POS]}\newline English translation: \textit{[target term]} \\

\midrule
\textsc{Parallel Sentences} \newline ($\leq$5 pairs) &
Retrieved parallel sentence pairs containing words similar to those in the source sentence. Two examples per source word are provided. &
To help with the translation, here is a translated sentence with words similar to ``\textit{[word]}'' in a list of translated \textit{[Language]}-English reference sentences:\newline \textit{[Language]} sentence: \textit{[source sentence]}\newline English translation: \textit{[target sentence]} \\

\midrule
\textsc{Grammar Passage} &
An excerpt from a descriptive grammar of the source language, covering relevant morphological or syntactic phenomena. &
To help with the translation, here is a passage retrieved from a \textit{[Language]} grammar book:\newline \textit{[grammar excerpt]} \\

\midrule
\textsc{Closing Instruction} &
Instructs the model to provide step-by-step metalinguistic reasoning and place the final translation in a \texttt{\textbackslash boxed\{\}} environment. &
Now write the translation of the sentence: \textit{[source sentence]}\newline Please provide your meta-linguistic reasoning step by step, and put your final translation within \texttt{\textbackslash boxed\{\}}. \\

\bottomrule
\end{tabular}
\end{table*}

\end{document}